\documentclass[journal]{IEEEtran}
\ifCLASSINFOpdf
\else
\fi
\usepackage[cmex10]{amsmath}
\usepackage{amssymb}
\usepackage[numbers]{natbib}
\usepackage{graphicx}
\usepackage{svg}
\begin{document}
%
\title{DeepStratNet: A Context-Aware Coordinate Regression Framework for Seismic Horizon Tracking under Sparse Labels}
%
%
%

%
%
%

\author{Aniq~Ahmad,
        Musham~Ahmad~Malik,
        Ahmad~Mustafa,~\IEEEmembership{Member,~IEEE,}
        and~Heather~Bedle%
\thanks{Aniq Ahmad and Musham Ahmad Malik contributed equally to this work.}%
\thanks{Aniq Ahmad and Heather Bedle are with Attribute Assisted Seismic Processing and Interpretation (AASPI), University of Oklahoma, USA (aniq.ahmad-1@ou.edu; hbedle@ou.edu).}%
\thanks{Musham Ahmad Malik is with the CIVIL Lab, Information Technology University, Lahore, Pakistan (musham656@gmail.com).}%
\thanks{Ahmad Mustafa is with the SDAIA Joint Research Center for AI (JRCAI), King Fahd University of Petroleum and Minerals, Dhahran, Saudi Arabia (ahmad.mustafa@kfupm.edu.sa).}%
\thanks{Corresponding author: Ahmad Mustafa (ahmad.mustafa@kfupm.edu.sa).}%
}
\markboth{Journal of \LaTeX\ Class Files,~Vol.~13, No.~9, September~2014}%
{Shell \MakeLowercase{\textit{et al.}}: Bare Demo of IEEEtran.cls for Journals}
%



\maketitle

\begin{abstract}

Automatic horizon tracking is a foundational task in 3D seismic interpretation. Most existing deep learning approaches formulate it as dense semantic segmentation, typically using U-Net-based architectures. The model produces a probability map over all pixels that must be post-processed to extract precise horizon coordinates, while horizon picks in time/depth must be converted into dense masks for training. Unpicked seismic traces are consequently treated as background, which can hinder convergence, and both pre- and post-processing can introduce errors into the final interpretation. Moreover, 2D segmentation models do not inherently capture inter-slice context, while 3D models are often computationally prohibitive. We instead formulate horizon tracking as a bounded coordinate regression problem, where the model directly predicts the time/depth coordinate of the target horizon at each lateral position. We propose a lightweight regression head compatible with any pretrained vision backbone, coupled with an LSTM module to model inter-slice context and produce a continuous horizon surface across the volume. A combination of L1 and L2 losses supervises predictions at valid horizon picks, while a geology-informed smoothness regularization enforces lateral continuity between successive traces. Under controlled experimental conditions, we evaluate four pretrained vision backbones—ResNet-50, ResNet-101, ViT-Large, and ConvNeXt-Large—under both segmentation and coordinate regression configurations on a seismic volume from New Zealand. The proposed approach consistently outperforms its segmentation counterparts quantitatively, using metrics including RMSE and PCC, and qualitatively, while also demonstrating greater robustness to increasing sparsity of training picks. Finally, we show that prediction variation across successive traces captures local variations in geological complexity, providing an automated quality control measure for downstream seismic interpretation.

\end{abstract}


%
\IEEEpeerreviewmaketitle

\section{Introduction}
%
%
%
\IEEEPARstart{G}{eophysical} data interpretation has been transformed by the rapid development of deep learning. It enabled significant improvements in the automation of seismic interpretation. Modern day deep learning models have demonstrated impressive performance across a diverse range of applications such as object classification, detection, and segmentation \cite{krizhevsky2017imagenet,li2017traffic,temel2018cure,zhao2019object}, medical image analysis for the diagnosis of medical conditions such as tumors and diabetes \cite{kumar2021brain,kokilepersaud2023clinically} and processing remote sensing data for crop monitoring \cite{sarma2022learning}, to name a few.

Deep learning models have been successfully applied to a wide range of subsurface seismic interpretation tasks such as fault detection \cite{mustafa2024visual}, sub-resolution fault mapping\cite{huang2026seismic}, salt body segmentation \cite{lubo2021exhaustive,ahmad2026domain}, and sink hole detection \cite{ahmad2025interactive}. They are also employed in the seismic inversion properties mapping \cite{pintea2022seismic}. 

Seismic horizon interpretation is particularly crucial for structure and reservoir modeling as well as for understanding the depositional history of a given area. Traditionally, horizons are picked semi-automatically as seed points manually created by an interpreter on a seismic reflection of interest. An autopicker tool is then used to extract horizon surfaces from the 3D volume by picking waveform signals exhibiting high similarity to those with the seed points. The process typically goes through several iterations to create the final extraction. 

In recent years, there has been increasing interest in using deep neural networks to automatically interpret horizons from seismic volumes given a few training sections containing labels for the horizon of interest \cite{yang2020seismic}. Machine learning approaches to horizon interpretation typically treat the problem as one of image segmentation whereby the goal is to infer all image pixels related to horizon as target of interest \cite{zhao2025seismic}. The 2D seismic section is provided as input to an image segmentation model such as the U-Net \cite{luo2022attention,liao2024deep}. The output of the model is a segmentation mask having the same dimensions as the input image. All pixels on and close to the object of interest are given the label ‘1’ whereas all other image pixels are labeled as ‘0’. This strategy has several drawbacks.

Firstly, segmentation models do not assume any prior constraints regarding the object being searched for. The assumption is that the target object can manifest itself in any pose, orientation, and number of instances. While this is an advantage to have for computer vision applications involving perception in self-driving cars, robots etc., it adds unnecessary complications in the modeling of seismic horizons which are mostly smooth, connected structures expected to occur at exactly one depth value per lateral position in a 2D seismic image. Secondly, seismic interpreters create horizon labels as seed points on seismic reflection data that needs conversion into label mask \cite{tschannen2020extracting} for training a dense segmentation model. This can potentially introduce an error into the training labels, the severity of which may depend on the spatial density of the initial picks and the specific label pre-processing scheme employed.

Thirdly, the segmentation model outputs its horizon predictions in the form of a label mask. Such models normally overestimate the vertical extent of the target horizon, predicting clouds of ‘1’s in its proximity. Additionally, depending on the image signal quality, it may also produce erroneous horizon predictions at considerable distances from the target horizon. Further processing is generally performed on such label masks to precisely identify the 3D coordinates of the target horizon on the seismic reflection image \cite{tschannen2020extracting,liao2025automatic}. This imprecision in the raw network output leads to another source of error in the final horizon interpretation.

Accordingly, prior works in this area use standard segmentation losses modified in various ways to account for the class imbalance caused by the sparse horizon picks processed into label masks for training deep networks\cite{tschannen2020extracting,peters2019multiresolution}. While these specialized loss functions can assist with stabilizing network convergence during training, the use of segmentation formulation means models predictions are automatically penalized at a pixel level and thus do not allow for incorporating smoothness-based geological priors preventing large variations in successive horizon picks predicted by the model. 

Additionally, many of the pretrained vision models released by the computer vision community for segmentation tasks were originally developed for static, standalone images. Seismic slices, however, whether in the inline or crossline direction, are not truly independent: each is a 2D view of a larger, spatially continuous subsurface configuration. Approaches that adapt these image-based models for horizon interpretation therefore have limited opportunity to exploit the spatial context shared across neighboring slices during training and inference. Incorporating this context is likely to be especially beneficial in structurally complex settings, where horizon continuity across slices carries important geological information.

In an attempt to address the problems with the segmentation methods outlined above, we propose a novel reformulation of the horizon-tracking application. Specifically, we set up horizon estimation as a bounded regression problem whereby given a 2D seismic reflection image, the output is a vector of estimated horizon depth values, one for each lateral position along the given image. By constraining the output to have exactly one point estimate per lateral position and by further restricting the point estimates to be bounded in range by the upper and lower time/depth values, respectively of the seismic image, we narrow down the space of possible solutions in an otherwise highly ill-posed problem.

Towards this end, we build a deep learning framework named DeepStratNet, a framework that reformulates horizon interpretation as a regression problem and jointly models the spatial and sequential geological information contained in a seismic volume. Our central contribution is a lightweight regression head that can be plugged in on top of any pretrained visual backbone. Because the head is backbone agnostic, it inherits the rich visual representations that deep networks learn from large scale natural image datasets and can be paired with a range of state of the art architectures. Our model employs a dedicated LSTM-based recurrent and attention module that comprises a window of neighboring inlines/crosslines, thereby capturing the continuity of subsurface structures across the volume. The attention module computes weights that quantify how much each slice in the context window contributes to the horizon prediction on the current slice, linking the extracted spatial features to Inline/crossline relationships. We train the network with a masked regression loss that penalizes the deviation between predictions and ground truth only over valid picks, so unpicked regions do not contribute spurious gradients. We further incorporate a geology-based loss that enforces consistency between successive picks and embeds a smoothness prior that discourages geologically implausible jumps between adjacent lateral positions. 

The primary contributions of this work are summarized as follows:

\begin{itemize}
\item \textbf{Bounded Regression Reformulation}: We address the structural limitations of dense segmentation formulations by re-framing horizon interpretation as a bounded 1D depth-coordinate regression problem. To realize this, we propose a lightweight, backbone-agnostic regression head that plugs in to any choice of the pretrained vision backbones to directly output a horizon depth values per lateral position in the image. The proposed module reduces the complexity of the solution space from predicting a dense probability map of size $H\times W$ to predicting a 1D vector of length $W$. Aside from cutting down the memory footprint and computational overhead, our approach eliminates the need for heavy spatial feature decoders and heuristic post-processing to extract precise depth/time coordinates for the target horizon.
\item \textbf{Spatial Context Modeling}: An LSTM module processes dense vision backbone embeddings for neighboring seismic slices on either side of the target image to produce spatially contextualized vector representations for the entire sequence. 
\item \textbf{Spatial Attention Weighting}: The spatial attention module computes learned weights to quantify the contribution of each neighboring slice to the target image's horizon prediction. The weighted aggregation of the LSTM embeddings across the entire sequence is passed through a final projection layer to produce the horizon coordinate vector for the target image.  
\item \textbf{Geology-Informed Loss Formulation}: We introduce a composite optimization objective comprising a masked regression loss to handle sparse ground-truth picks without gradient noise from unlabeled regions, and a geology-inspired smoothness loss that penalizes geologically implausible depth jumps across adjacent lateral positions.
\end{itemize}

\section{Related Work}
\label{sec:related_work}

Deep learning approaches for seismic horizon tracking fall under three primary paradigms: discrete patch/trace classification, semantic image/volumetric segmentation, and probabilistic density regression. Early convolutional frameworks formulate horizon picking as a 2D image patch classification task. For example, Yang and Sun (2020)~\cite{yang2020seismic} extract small 2D seismic amplitude patches surrounding target samples and train a convolutional neural network (CNN) to classify whether the central pixel corresponds to a horizon or non-horizon class. 

To overcome the spatial limitations of patch-based methods, subsequent works reframe horizon interpretation as pixel-wise or voxel-wise semantic segmentation. Wu and Zhang (2018) model horizon tracking as 2D trace segmentation, partitioning seismic cross-sections into distinct stratigraphic zones bounded by horizon surfaces \cite{wu2018deep}. Extending segmentation to three dimensions, Tschannen et al. (2020) and Liao et al. (2025) frame picking as 3D volumetric segmentation, predicting horizon class probabilities for every voxel within 3D seismic subvolumes \cite{tschannen2020extracting, liao2025automatic}. Similarly, Zhao and Zhao (2025) utilize semantic segmentation to establish direct mapping relationships between 3D seismic volumes and single or multiple horizon boundaries \cite{zhao2025seismic}.

Rather than enforcing rigid class assignments, a third class of methods models horizon picking through continuous probability estimation. Peters et al. (2018) reformulate horizon tracking as a conditional density estimation problem that outputs continuous horizon probability maps across full seismic sections \cite{peters2019multiresolution}. Building upon probabilistic framing, Liao et al. (2024) conceptualize horizon detection as conditional probability density estimation along individual seismic traces, predicting a smooth vertical probability distribution whose peak location indicates the horizon boundary \cite{liao2024deep}.

The approaches described above all require horizon labels to be preprocessed into a form compatible with the input-output framework employed. This potentially introduces a certain degree of error into the process. Under our proposed framework in contrast, the neural network predicts a horizon coordinate for each lateral position in the image. Consequently, we can directly compare the output of the model to the ground-truth horizon picks produced by the interpreter without having to convert them first into intermediate forms such as segmentation masks, probability distributions etc.

The neural network architectures applied to horizon tracking have evolved from basic convolutional structures to advanced multi-scale, residual, and transformer-based models. Initial classification approaches rely on standard sequential 2D CNNs consisting of alternating convolutional layers, activation functions, max-pooling layers, and fully connected layers terminating in a classification head \cite{yang2020seismic}. To process multi-scale seismic features while maintaining full spatial resolution, multi-scale encoder-decoder networks have become widely adopted. Wu and Zhang (2018) utilize a SegNet-style convolutional encoder-decoder architecture, utilizing the encoder to extract hierarchical amplitude features and the decoder with upsampling layers to restore trace resolution \cite{wu2018deep}. Peters et al. (2018) and Liao et al. (2024) implement multi-resolution U-Net variants, leveraging skip connections to combine fine-grained spatial details from shallow encoder layers with broad semantic context from deep decoder layers \cite{peters2019multiresolution, liao2024deep}. To better capture long-range spatial dependencies across complex geological structures, recent architectures incorporate self-attention mechanisms. Zhao and Zhao (2025) introduce the TransUnet architecture \cite{zhao2025seismic}, integrating vision transformer modules into a U-Net encoder-decoder structure to combine local convolutional feature extraction with global attention capabilities. Moving to purely volumetric self-attention, Liao et al. (2025) implement a 3D volumetric transformer network utilizing 3D shifted-window self-attention blocks and 3D patch-expanding layers to capture global 3D spatial continuity across seismic subvolumes \cite{liao2025automatic}. 

Our contribution in contrast is not any one specific architecture but rather a neural network module that can be plugged on top of any pretrained neural network backbone to directly predict horizon coordinates for the given seismic image. This way, we are able to leverage the rich representations learned by large foundation models released by the compute vision community such as the ViT, ConvNeXt \cite{dosovitskiy2021an, liu2022convnet} etc., while being able to adapt them for out-of-domain task of seismic horizon tracking. Simultaneously, we are able to model and learn the spatial context across multiple neighboring slices to predict the horizon coordinates for the target image with greater accuracy while keeping the computational cost lower than 3D approaches.     

To enable training from sparse annotations, several studies employ masked or projected loss functions. Peters et al. (2018) propose a projected loss function that evaluates prediction errors strictly at sparsely labeled pixel locations while passing full, uncropped seismic sections through the network \cite{peters2019multiresolution}. Similarly, Tschannen et al. (2020) and Liao et al. (2025) apply a masked cross-entropy loss function, evaluating gradients exclusively on a few annotated 2D slices per 3D subvolume while ignoring unannotated voxels \cite{tschannen2020extracting, liao2025automatic}. Liao et al. (2024) likewise utilize a masked loss controlled by seed-point intersections along intersecting profile lines \cite{liao2024deep}. To account for manual picking ambiguity and reduce label imbalance, target labels are frequently smoothed. Peters et al. (2018), Liao et al. (2024), and Liao et al. (2025) encode discrete horizon picks by convolving target positions in the vertical/time direction with a 1D Gaussian kernel, generating smooth probability target distributions \cite{peters2019multiresolution, liao2024deep, liao2025automatic}. 

One limitation with all these methods is the fact that they produce a dense output that has to be post processed to extract precise horizon coordinates that can potentially introduce an error. On the other hand, models trained under our proposed framework directly output horizon coordinates for the given slice. Additionally, this allows us to impose geological consistency based regularization over the loss function. This would not have been possible in dense output based segmentation/regression frameworks.

\section{Methodology}
\label{sec:methodology}

\subsection{Problem Formulation and Dataset Representation}
\label{subsec:problem_formulation}

We model a three-dimensional (3D) seismic volume as a sequence of two-dimensional (2D) images lying along a specific spatial direction. This is the inline coordinate axis in our case. 

Let the dataset be defined as $\mathcal{V} = \{(\mathbf{X}_i, \mathbf{y}_i)\}_{i=1}^{N_v}$, consisting of $N_v$ paired training samples, where $\mathbf{X}_i$ represents the input seismic sequence and $\mathbf{y}_i$ denotes the corresponding target horizon interpretation. Rather than evaluating a target inline in isolation, each input sample $\mathbf{X}_i$ is represented as a spatio-temporal sequence of $T$ adjacent seismic inlines centered on the target inline index $i$:

\begin{equation}
\mathbf{X}_i = \left[ \mathbf{S}_{i-k}, \dots, \mathbf{S}_i, \dots, \mathbf{S}_{i+k} \right] \in \mathbb{R}^{T \times C \times H \times W},
\label{eq:input_tensor}
\end{equation}

where $T = 2k + 1$ denotes the context window size ($k \ge 1$), $C$ represents the channel dimension, and $H$ and $W$ correspond to the standardized depth/time vertical resolution and crossline horizontal resolution, respectively. The central slice $\mathbf{S}_i$ represents the target evaluation inline, while neighboring slices $\mathbf{S}_{t \neq i}$ supply spatial contextual constraints.

The target horizon interpretation corresponding to sample $i$ is formulated as a 1D vector of continuous depth coordinates:

\begin{equation}
\mathbf{y}_i = \left[ y_{i,1}, y_{i,2}, \dots, y_{i,W} \right]^\top \in [0, 1]^W,
\label{eq:target_vector}
\end{equation}

where $y_{i,j}$ defines the normalized vertical horizon coordinate at crossline position $j \in \{1, \dots, W\}$. Columns lacking valid horizon interpretations are masked out during optimization.
\subsection{DeepStratNet Architecture}
\label{subsec:architecture}

The model processes the input sequence $\mathbf{X}_i$ through four main stages: spatial feature extraction, temporal sequence modeling, temporal attention aggregation, and horizon regression. Figure~\ref{fig:deepstratnet_methodology} illustrates the end-to-end processing pipeline.

\subsubsection{Spatial Feature Extraction and Global Pooling}
Each seismic slice $\mathbf{S}_t \in \mathbb{R}^{C \times H \times W}$ within the sequence ($t \in \{1, \dots, T\}$) is independently processed through a parameterized 2D convolutional or vision transformer backbone $\mathcal{B}(\cdot)$ to yield spatial feature maps:

\begin{equation}
\mathbf{F}'_t = \mathcal{B}(\mathbf{S}_t) \in \mathbb{R}^{C' \times H' \times W'},
\label{eq:backbone_extraction}
\end{equation}

where $C'$ represents the latent feature channels, and $H', W'$ denote the spatial dimensions of the bottleneck representation. A Global Average Pooling ($\mathrm{GAP}$) operation is applied across the spatial dimensions of $\mathbf{F}'_t$ to condense each slice into a 1D vector of embedding dimension $C'$:

\begin{equation}
\mathbf{f}_t = \mathrm{GAP}(\mathbf{F}'_t) \in \mathbb{R}^{C'},
\label{eq:gap_vector}
\end{equation}

 Cascading these vectors chronologically across the sequence length $T$ forms the latent sequence:

\begin{equation}
\mathbf{F} = [\mathbf{f}_1, \mathbf{f}_2 \dots  \mathbf{f}_T] \in \mathbb{R}^{C' \times T},
\label{eq:feature_sequence_matrix}
\end{equation}

where $\mathbf{f}_t \in \mathbb{R}^{C'}, \; \forall t \in \{1, \dots, T\}$.

\subsubsection{Spatial Sequence Modeling}
To model structural variations across adjacent inlines, the feature sequence $\mathbf{F}$ is processed by a Recurrent Neural Network (LSTM):

\begin{equation}
\mathbf{H} = \mathrm{LSTM}(\mathbf{F}), ,
\label{eq:lstm_sequence}
\end{equation}

where $\mathbf{H} = \left[ \mathbf{h}_1, \mathbf{h}_2, \dots, \mathbf{h}_T \right]^\top \in \mathbb{R}^{T \times d_h}$ and $\mathbf{h}_t \in \mathbb{R}^{d_h}$ denotes the temporal hidden state at index $t$ with hidden dimensionality $d_h$.

\begin{figure*}[t]
    \centering
    \includegraphics[width=1\linewidth]{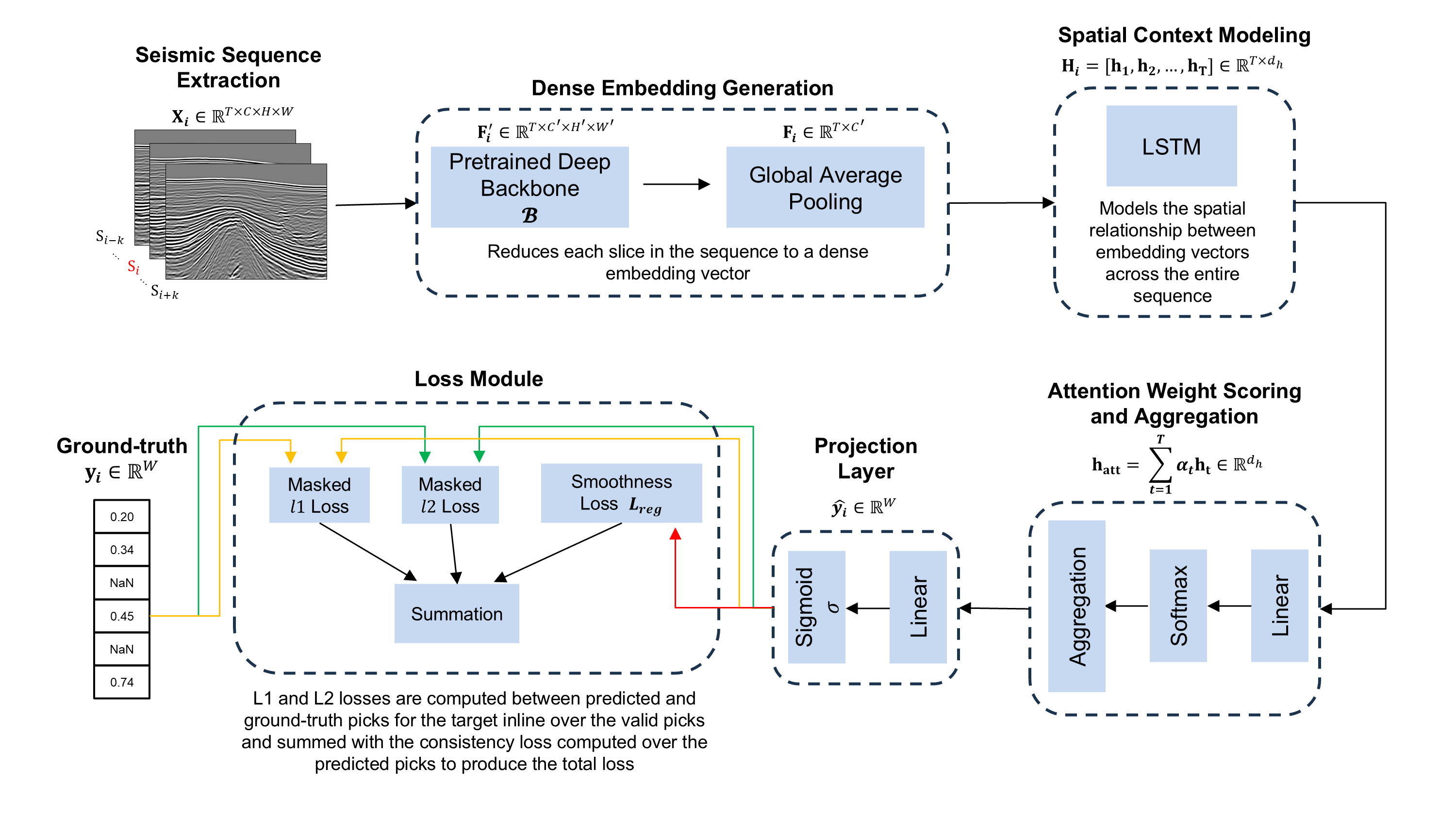}
    \caption{Architectural pipeline of the proposed seismic horizon picking framework. Given a sequence of seismic slices $X_i$, a pretrained backbone extracts slice-level dense embeddings $F_i$. An LSTM module models sequential spatial context across slices, followed by an attention-based aggregation layer that combines feature representations into $h_{\text{att}}$. A projection layer outputs the target pick prediction $\hat{y}_i \in \mathbb{R}^W$. The model is trained using a composite loss function comprising masked $l_1/l_2$ reconstruction losses and a smoothness regularization term $L_{\text{reg}}$.}
    \label{fig:deepstratnet_methodology}
\end{figure*}





\subsubsection{Spatial Attention Mechanism}
To account for varying relevance and noise levels across adjacent seismic slices, a soft attention mechanism calculates a scalar importance score $e_t$ for each hidden state $\mathbf{h}_t$:

\begin{equation}
e_t = \mathbf{w}_{\mathrm{att}}^\top \mathbf{h}_t + b_{\mathrm{att}},
\label{eq:attention_score_ref}
\end{equation}

where $\mathbf{w}_{\mathrm{att}} \in \mathbb{R}^{d_h}$ and $b_{\mathrm{att}} \in \mathbb{R}$ are learnable parameters. The scalar scores are normalized via a Softmax function to obtain attention weights $\alpha_t$:

\begin{equation}
\alpha_t = \frac{\exp(e_t)}{\sum_{j=1}^{T} \exp(e_j)}, \quad \text{such that} \quad \sum_{t=1}^{T} \alpha_t = 1.
\label{eq:softmax_attention}
\end{equation}

The final context vector $\mathbf{h}_{\mathrm{att}}$ is computed as the weighted sum of the hidden states:

\begin{equation}
\mathbf{h}_{\mathrm{att}} = \sum_{t=1}^{T} \alpha_t \mathbf{h}_t \in \mathbb{R}^{d_h}.
\label{eq:attention_context}
\end{equation}

\subsubsection{Horizon Prediction Head}
The aggregated context representation $\mathbf{h}_{\mathrm{att}}$ is mapped to the final normalized horizon coordinates $\hat{\mathbf{y}} \in [0, 1]^W$ using a fully connected projection layer followed by a Sigmoid activation:

\begin{equation}
\hat{\mathbf{y}} = \sigma \left( \mathbf{W}_o \mathbf{h}_{\mathrm{att}} + \mathbf{b}_o \right),
\label{eq:final_prediction}
\end{equation}

where $\mathbf{W}_o \in \mathbb{R}^{W \times d_h}$, $\mathbf{b}_o \in \mathbb{R}^W$, and $\sigma(\cdot)$ represents the element-wise logistic sigmoid function.

\subsection{Loss Functions}
\label{subsec:loss_functions}

The network is optimized using a composite objective function combining pointwise spatial distance metrics and a structural crossline smoothness constraint.

\subsubsection{Pointwise Regression Loss}
The discrepancy between predicted horizon values $\hat{\mathbf{y}}$ and ground-truth interpretations $\mathbf{y}$ is measured using a combined Mean Squared Error ($\mathrm{MSE}$) and Mean Absolute Error ($\mathrm{MAE}$) loss over valid crossline indices $\Omega$:

\begin{equation}
\mathcal{L}_{\mathrm{reg}}(\hat{\mathbf{y}}, \mathbf{y}) = \frac{1}{|\Omega|} \sum_{j \in \Omega} \left( (\hat{y}_j - y_j)^2 + |\hat{y}_j - y_j| \right).
\label{eq:pointwise_loss}
\end{equation}

\subsubsection{Crossline Smoothness Regularization}
To prevent unphysical vertical oscillations along adjacent crossline positions, a first-order difference smoothness loss is enforced across the predicted horizon curve:

\begin{equation}
\mathcal{L}_{\mathrm{smooth}}(\hat{\mathbf{y}}) = \frac{1}{W-1} \sum_{j=1}^{W-1} \left( \hat{y}_{j+1} - \hat{y}_j \right)^2.
\label{eq:smoothness_loss_ref}
\end{equation}

\subsubsection{Total Objective Function}
The total joint loss function $\mathcal{L}$ is defined as:

\begin{equation}
\mathcal{L} = \frac{1}{N_{v}}\sum_{1}^{N_v}\mathcal{L}_{\mathrm{reg}}(\hat{\mathbf{y}}, \mathbf{y}) + \lambda \mathcal{L}_{\mathrm{smooth}}(\hat{\mathbf{y}}),
\label{eq:total_loss_ref}
\end{equation}

where $\lambda > 0$ is a scalar hyperparameter balancing pointwise fidelity with spatial structural continuity and $N_{v}$ is the total number of labeled seismic inlines available for training.

\subsection{Evaluation Metrics}

The model is evaluated using only valid annotated horizon columns. The
reported metrics are root mean squared error (RMSE), Pearson correlation
coefficient (PCC), and coefficient of determination ($R^2$).

The root mean squared error is calculated as

\begin{equation}
\operatorname{RMSE} =
\sqrt{
\frac{1}{N}
\sum_{n=1}^{N}
(\hat{y}_n-y_n)^2
}.
\label{eq:rmse}
\end{equation}

The coefficient of determination is calculated as

\begin{equation}
R^2 =
1-
\frac{\displaystyle\sum_n(y_n-\hat{y}_n)^2}
{\displaystyle\sum_n(y_n-\bar{y})^2}.
\label{eq:r2}
\end{equation}

The Pearson correlation coefficient is calculated as

\begin{equation}
\operatorname{PCC} =
\frac{
\displaystyle\sum_n
(y_n-\bar{y})
(\hat{y}_n-\overline{\hat{y}})
}{
\sqrt{\displaystyle\sum_n(y_n-\bar{y})^2}
\sqrt{\displaystyle\sum_n
(\hat{y}_n-\overline{\hat{y}})^2}
}.
\label{eq:pcc}
\end{equation}

The predictions and reference horizons are normalized to the range $[0,1]$ during evaluation. The RMSE is first converted to pixel units by multiplying it by the image height of 224 pixels. Because the original seismic sections have 176 vertical samples with a sampling interval of 4 ms and are resized to 224 pixels for model input, the pixel based RMSE is subsequently converted to time units using the original sampling interval and the ratio between the original and resized image heights. Therefore, the RMSE reported in the results tables is expressed in milliseconds, whereas PCC and $R^2$ are dimensionless and remain unchanged by this scaling. Metrics are reported for both the training and test sets and are averaged across individual seismic slices.






\section{Results}


\begin{figure*}[t]
\centering
\includegraphics[width=\textwidth]{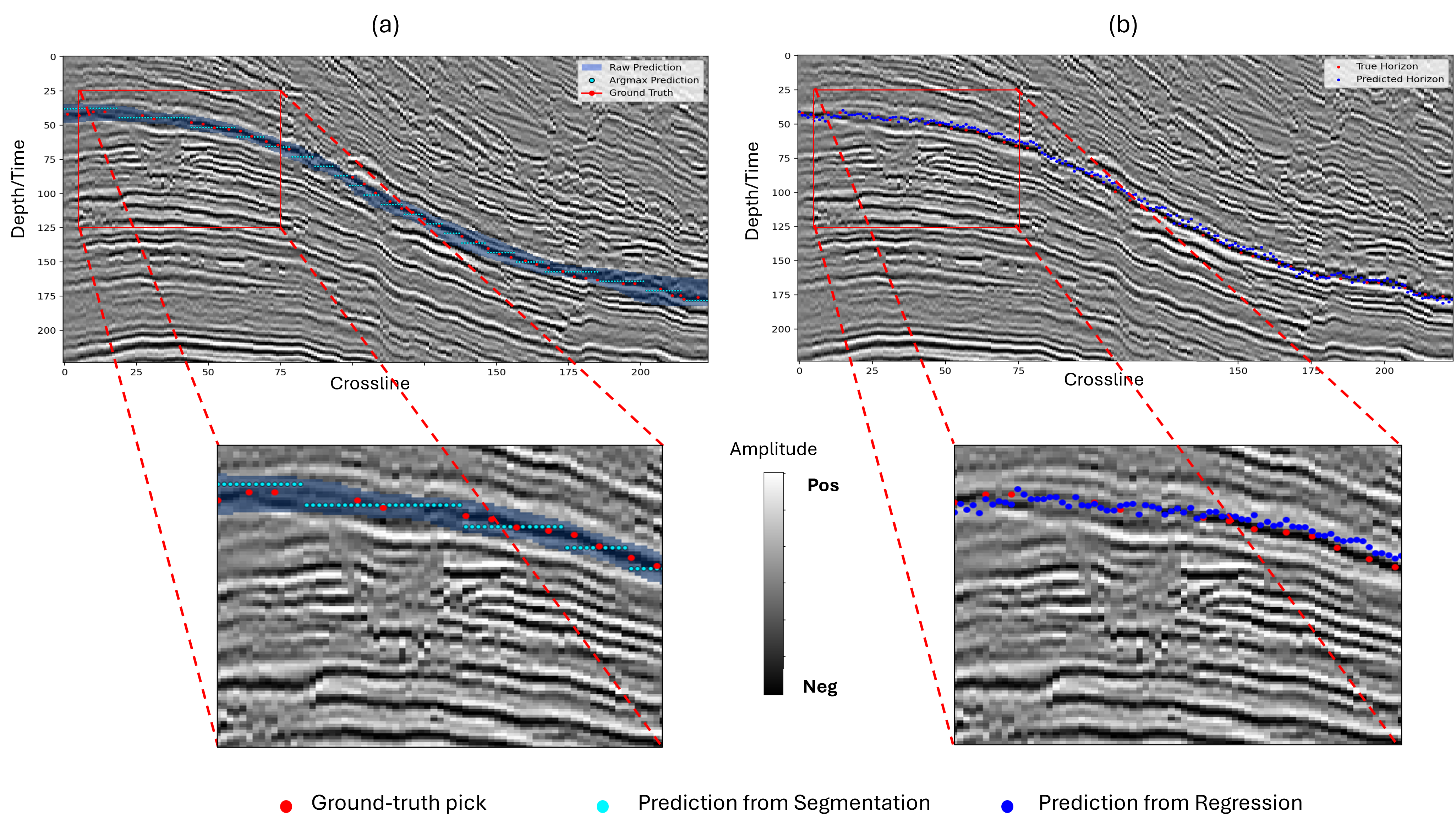}
\caption{Comparison of segmentation and regression model on an inline section. (a) shows the segmented mask and the horizon extracted using the argmax of the predicted logits. Extracting the horizon from the segmentation mask introduces additional error and uncertainty, as the extracted horizon does not consistently follow the reflection character. (b) shows the horizon directly predicted by the regression model, which accurately follows the reflection character of the target horizon.}
\label{fig:reg_vs_seg_il265}
\end{figure*}

\begin{figure}
    \centering
    \includegraphics[width=0.5\linewidth]{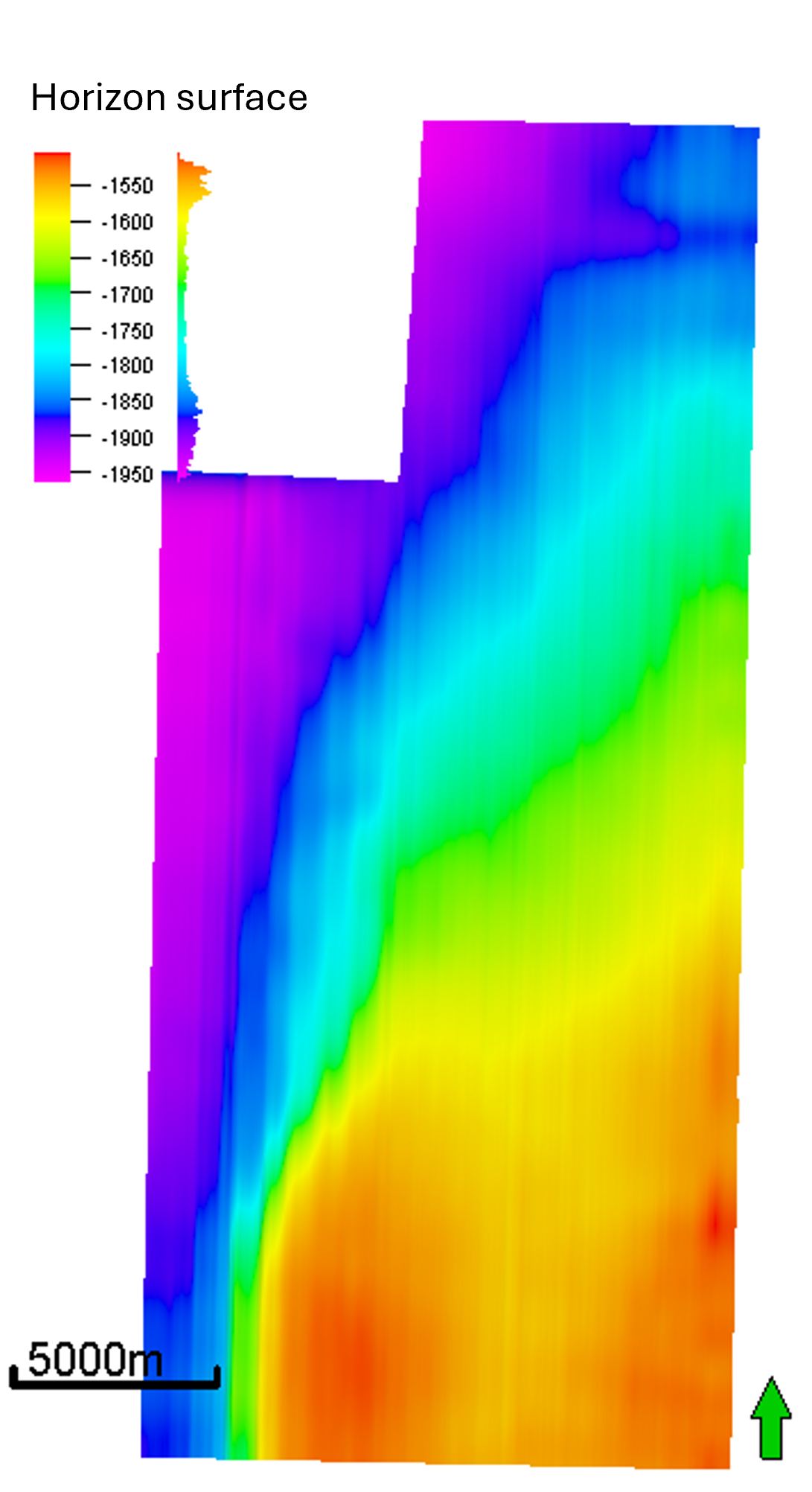}
    \caption{Time surface of the horizon generated from the model. The vertical axis shows two-way travel time (TWT) in milliseconds. Travel times increase toward the northwest, indicating that the horizon dips to the NW }
    \label{fig:horizon_surface}
\end{figure}

\begin{figure*}[t]
\centering
\includegraphics[width=\textwidth]{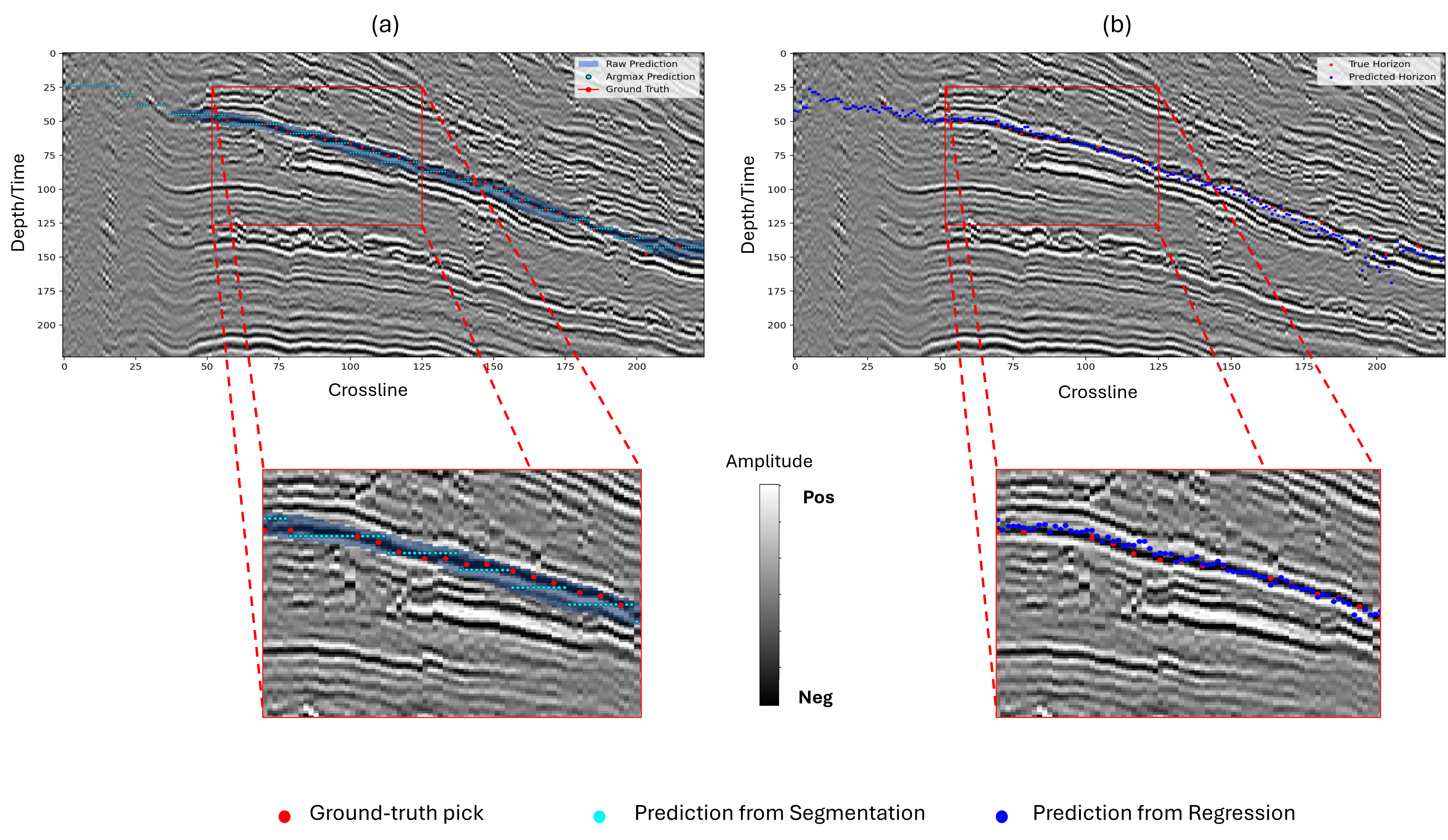}
\caption{Another Inline showing the comparison of segmentation and regression model.(a) and (b) shows the predicted shown from segmentation and regression model respectively. (b) also shows the irregular horizon picks predicted by the model. This character is noticed on couple of Inlines and it marks the presence of the channel.}
\label{fig:reg_vs_seg_il493}
\end{figure*}

\subsection{Dataset}

We performed an initial interpretation with the intention of mapping a deep-water meandering channel system on Maui 3D seismic volume from the Taranaki Basin, New Zealand. The variable amplitudes and chaotic reflection patterns make the reflector hard to follow. We picked a horizon using a 2D guided tracker. The sampling interval is 4 ms and the dominant frequency is 45 Hz. The dataset consists of a total of 572 inlines, of which we were able to trace the horizon on 22 inlines (uniformly spread over the volume). We used the 22 fully labeled sections as our training dataset. We selected 32 other inlines to serve as the held-out validation/test set, which included tie point lines and medium-picked sections. Each input slice was resized to a resolution of 224 x 224 pixels. Datasets were normalized per using z-score normalization to ensure consistent amplitude scaling.

\subsection{Backbones}

To demonstrate that our proposed module works for any choice of the pretrained backbone, we evaluated it in conjunction with four different backbone architectures: (1) ResNet-50, (2) ResNet-101 \cite{he2016deep}, (3) ViT-Large \cite{dosovitskiy2021an}, and (4) ConvNeXt-Large \cite{liu2022convnet}. We used the pretrained checkpoints provided by the official PyTorch library. Each pretrained backbone was finetuned and tested on the aforementioned dataset under identical conditions first with a segmentation head and thereafter with the proposed module. For the segmentation configuration, ResNet-50 and ResNet-101 were used as the backbones of the popular DeepLabV3 \cite{chen2017rethinking} architecture. ViT-Large and ConvNeXt-Large were adapted for dense prediction by adding convolutional projection and upsampling layers to their respective feature representations. To maintain consistency across the backbone architectures, the same Atrous Spatial Pyramid Pooling (ASPP) module, following the ASPP design used in DeepLab, was used with both ViT-Large and ConvNeXt-Large. The resulting segmentation outputs were then upsampled to the spatial dimensions of the input seismic image.

Training labels for the segmentation experiments were created by generating binary masks for each training inline denoting horizon picks with 1s and background pixels wth 0s. Thereafter, we convolved a gaussian kernel with the binary masks. This was followed by a thresholding operation to convert float values back to the binary domain. The purpose of this operation was to thicken the horizon labels for stable training with the segmentation loss. We used a standard weighted binary cross entropy loss function to train each of our segmentation backbones. Since the output of these models was a dense logit map, we extracted horizon picks by identifying the pixel with the maximum logit value along each vertical seismic trace along the test image. This is described by the equation 

\begin{equation}\
\hat{\mathbf{y}}_{seg}(x) = \operatorname*{argmax}_{y} \mathbf{S}(y, x),
\label{eq:seg_predict}
\end{equation}
where $\mathbf{S} \in \mathbb{R}^{H \times W}$ represents the predicted dense logit map, with $y \in \{1, \dots, H\}$ indexing the vertical depth samples and $x \in \{1, \dots, W\}$ indexing the seismic trace columns. In contrast to this standard segmentation pipeline, our proposed module eliminates the need for heuristic post-processing and label-thickening operations by directly formulating horizon identification as a continuous coordinate regression problem. 

To further test if the LSTM module was actually assisting the proposed module, we carried out an additional experiment where we dropped the LSTM from the proposed module, using only the target seismic inline to predict the corresponding horizon picks. The rest of the decoder structure and the training regimen remained the same as in Fig. \ref{fig:deepstratnet_methodology}. For brevity, we will subsequently refer to these configurations as (1) segmentation, (2) regression, and (3) regression - no LSTM.

Table \ref{tab:combined_results} shows the average training and test set performance of the various neural network backbones in the three aforementioned experimental configurations in terms of RMSE, PCC and $r^{2}$ coefficient, respectively. Horizon predictions generated by the proposed method as well its segmentation counterpart for the highest performing ConvNeXt backbone is shown for two different test slices in Fig. \ref{fig:reg_vs_seg_il265} and Fig. \ref{fig:reg_vs_seg_il493}, respectively. The continous horizon surface is shown in Fig. \ref{fig:horizon_surface}.

\subsection{Sparsity Analysis}

\begin{figure*}[t]
\centering
\includegraphics[width=\textwidth]{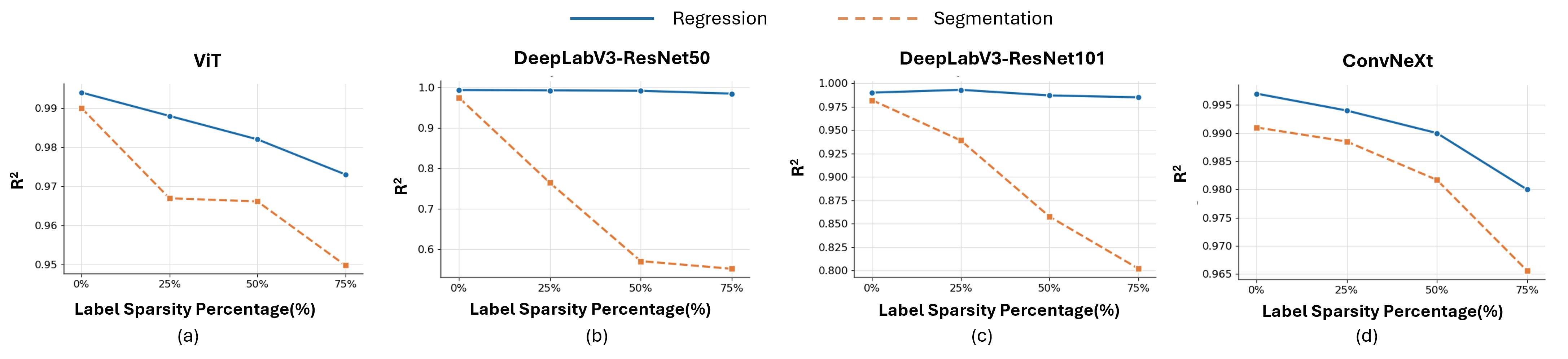}
\caption{Different backbones (a–d) showing the degradation in regression and segmentation model performance as the proportion of training labels decreases. Segmentation models show the fastest degradation across all backbones, with ResNet-50 and ResNet-101 (b–c) showing the largest performance drop, followed by ViT-Large (a) and ConvNeXt-Large (d). Regression models show a more consistent trend with a smaller drop as the number of training labels decreases across all backbones. }
\label{fig:sparsity}
\end{figure*}

We evaluated the effects of reducing training pick density on model performance by randomly deleting available picks on each training section at three different levels: 25\%, 50\%, and 75\% respectively. Each of the aforementioned model backbones were thereafter trained in both the segmentation and the proposed regression configurations on these reduced pick density datasets using the same training setup as before. Evaluation was carried out over the hold-out test-set. The average test-set $r^{2}$ coefficient at each pick density level for all backbones is plotted in Fig. \ref{fig:sparsity}.

\subsection{Uncertainty Analysis}
\begin{figure*}[t]
\centering
\includegraphics[width=\textwidth]{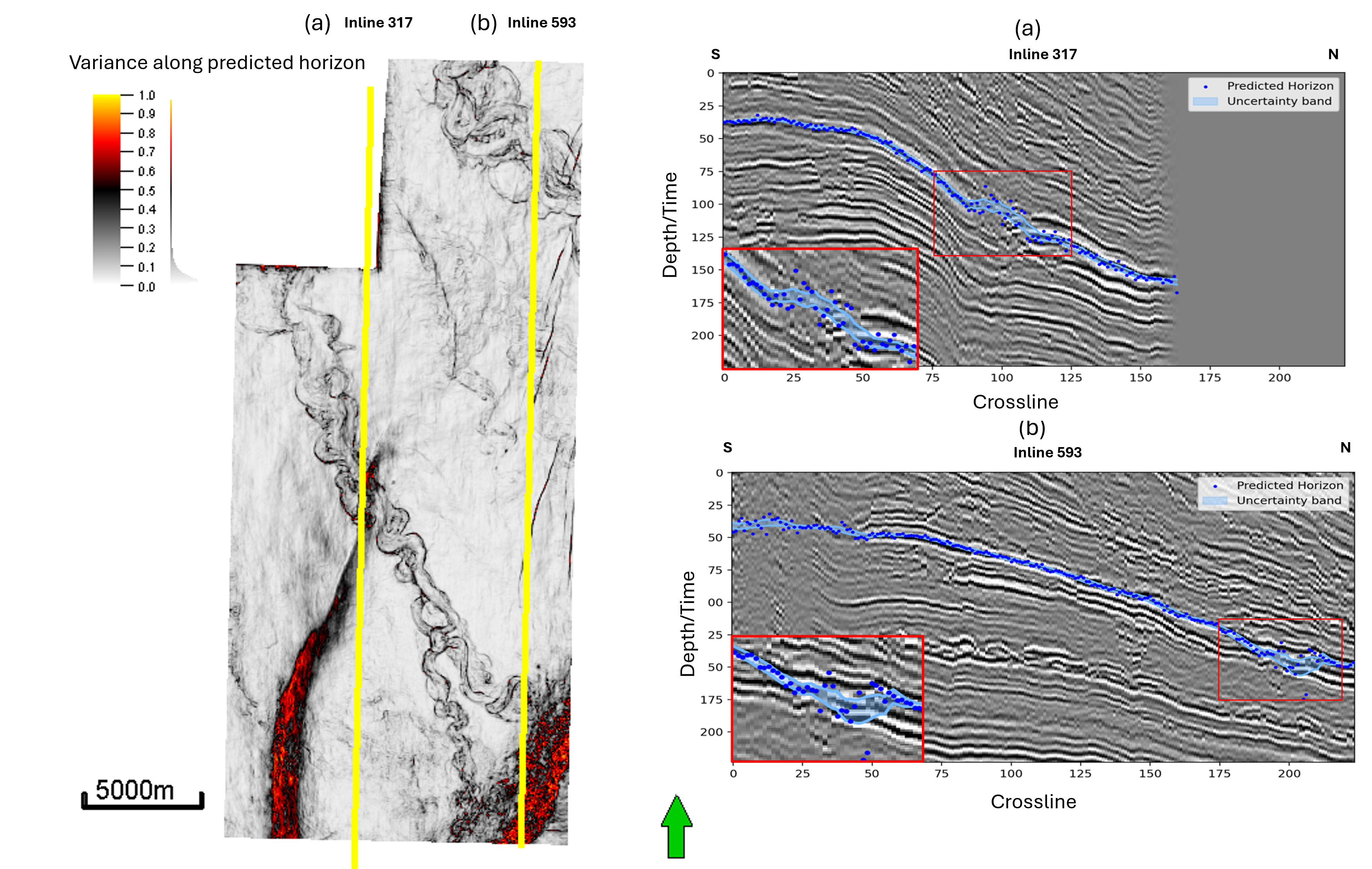}
\caption{Model uncertainty shown by the inter-point variance. A thin uncertainty band indicates low variance, whereas a thick uncertainty band indicates high variance. (a) and (b) show two examples of the predicted horizon, with each example including an individual view and a zoomed-in view. The high inter point variance regions coincide with chaotic amplitude patterns that indicate the presence of channel features which is confirmed by the seismic attribute (Variance) along the predicted horizon.}
\label{fig:uncertainty}
\end{figure*}

Since the models trained in the proposed regression configuration directly output horizon coordinates against a given input seismic section, we are able to calculate the uncertainty in model predictions for each horizontal position along the image. The uncertainty plot is based on the local variation of the predicted horizon along the crossline direction. A moving-window variance is
calculated as

\begin{equation}
v_j =
\operatorname{Var}
\left(
\hat{y}_{j-r},\ldots,\hat{y}_{j+r}
\right),
\label{eq:uncertainty_variance}
\end{equation}

where $r$ controls the size of the local window. The square root of this
variance is smoothed and displayed as an uncertainty band around the
predicted horizon. A wider band indicates greater local variation in the
prediction. Uncertainty visualizations on two independent test sections are shown in Fig. \ref{fig:uncertainty} for the best performing ConvNeXt backbone as per Table \ref{tab:combined_results}.

\begin{table*}[t]
\centering
\caption{Benchmark for various vision backbones under the Segmentation and Regression Configurations}
\label{tab:combined_results}
\renewcommand{\arraystretch}{1.3}
\begin{tabular}{lccc|ccc}
\hline
 & \multicolumn{3}{c|}{\textbf{Training Set}} & \multicolumn{3}{c}{\textbf{Test Set}} \\
\cline{2-7}
\textbf{Model} & \textbf{RMSE} & \textbf{PCC} & \textbf{$R^2$} & \textbf{RMSE} & \textbf{PCC} & \textbf{$R^2$} \\
\hline
DeepLabV3-ResNet50 (segmentation) & 10.78 ms & 0.9924 & 0.9733 & 13.36 ms & 0.9915 & 0.9753 \\
DeepLabV3-ResNet50 (regression - no LSTM) & 16.75 ms & 0.9826 & 0.9329 & 14.52 ms & 0.9968 & 0.9848 \\
\textbf{DeepLabV3-ResNet50 (regression)} & \textbf{10.25 ms} & \textbf{0.9969} & \textbf{0.9869} & \textbf{8.77 ms} & \textbf{0.9985} & \textbf{0.9943} \\
\hline
DeepLabV3-ResNet101 (segmentation) & 9.24 ms & 0.9940 & 0.9789 & 12.16 ms & 0.9914 & 0.9820 \\
DeepLabV3-ResNet101 (regression - no LSTM) & 19.30 ms & 0.9873 & 0.9353 & 18.35 ms & 0.9967 & 0.9761 \\
\textbf{DeepLabV3-ResNet101 (regression)} & \textbf{11.03 ms} & \textbf{0.9959} & \textbf{0.9806} & \textbf{11.56 ms} & \textbf{0.9978} & \textbf{0.9906} \\
\hline
ConvNeXt (segmentation) & 10.97 ms & 0.9930 & 0.9656 & 10.37 ms & 0.9965 & 0.9916 \\
ConvNeXt (regression - no LSTM) & 7.10 ms & 0.9963 & 0.9907 & 10.43 ms & 0.9982 & 0.9917 \\
\textbf{ConvNeXt (regression)} & \textbf{7.32 ms} & \textbf{0.9976} & \textbf{0.9941} & \textbf{8.32 ms} & \textbf{0.9989} & \textbf{0.9971} \\
\hline
ViT (segmentation) & 9.84 ms & 0.9940 & 0.9759 & 10.12 ms & 0.9966 & 0.9920 \\
ViT (regression - no LSTM) & 8.05 ms & 0.9894 & 0.9751 & 14.90 ms & 0.9968 & 0.9845 \\
\textbf{ViT (regression)} & \textbf{5.21 ms} & \textbf{0.9978} & \textbf{0.9950} & \textbf{8.96 ms} & \textbf{0.9985} & \textbf{0.9946} \\
\hline
\end{tabular}
\end{table*}









\section{Discussion}

The results indicate that direct coordinate regression changes the horizon tracking problem from dense pixel classification to direct estimation of the horizon position along each seismic trace. The consistent behavior across different feature extraction architectures suggests that the observed improvement is not tied to a particular backbone, but instead arises from the formulation of the prediction task itself. By estimating a single horizon coordinate for each trace, the model avoids the intermediate conversion of a dense probability map into a horizon pick. This is particularly important for seismic interpretation, where the final objective is the location and continuity of the reflector rather than a pixelwise classification map. The qualitative examples in Fig. \ref{fig:reg_vs_seg_il265} and Fig. \ref{fig:reg_vs_seg_il493} further suggest that this formulation better preserves the continuous character of seismic reflections, reduces the discretization effects introduced when low resolution feature representations are converted back to image resolution and produces a continuous horizon surface as shown in \ref{fig:horizon_surface}. The horizon surface shows the continuous variation of geological layers across the seismic survey that is dipping in the NW direction.

An important implication of this formulation is the role of spatial context in horizon tracking. A horizon is a single value per trace and multiple continuous time values across Inlines/crosslines in a seismic volume. Its continuous nature is guided by local geology and rock deposition. The improvement obtained by incorporating the LSTM therefore reflects the importance of modeling this spatial dependency rather than simply increasing the complexity of the feature extractor. The recurrent component provides information from neighboring sections that can help constrain predictions where the seismic response is locally ambiguous. This is particularly relevant in 3D seismic interpretation, where reflector continuity extends beyond the information available in an individual inline or crossline.

The behavior under sparse labeling provides another important distinction between regression and segmentation based formulations. In practical seismic interpretation, complete horizon labeling is often unrealistic and interpreters commonly provide sparse picks that follow the most reliable portions of a reflector. The regression formulation naturally accommodates this workflow because supervision is applied only where a horizon coordinate is available. Segmentation, in contrast, requires an explicit distinction between horizon and background across the image. As the density of labeled traces decreases the assumption that all remaining locations represent background becomes increasingly inconsistent with the actual interpretation task. The degradation observed in Fig .~\ref{fig:sparsity} for segmentation under sparse labeling therefore highlights a fundamental mismatch between dense classification and the way horizons are commonly interpreted. The regression formulation is better aligned with the available supervision because an unlabeled trace remains simply unlabeled rather than being treated as evidence against the horizon.

The geology based smoothness constraint (Eq.~\ref{eq:smoothness_loss_ref}) further provides an important connection between the learning objective and the physical structure of the seismic volume. Rather than relying entirely on the training labels to determine how predictions should behave between sparse observations, the regularization encourages spatially coherent horizon trajectories. This does not impose a fixed geometric shape on the horizon. Instead, it discourages unrealistic local variations while allowing the model to follow changes in reflector geometry. This is useful when labeled control is limited, although its effectiveness depends on the assumption that the target horizon remains locally continuous.

The uncertainty analysis depicted in Fig.~\ref{fig:uncertainty} shows that the variance in the model's outputs correlates with the local geological complexity. In areas with high signal-to-noise ratios and continuous seismic reflectors, the model exhibits low predictive uncertainty, reflected in a tight vertical spread around the pick. Conversely, when structural continuity breaks down such as across complex channel cuts-the vertical uncertainty spread widens proportionally. This direct coupling between model confidence and geological clarity provides a mechanism for automated Quality Control (QC), allowing geoscientists to automatically flag ambiguous structural zones for expert review prior to downstream seismic interpretation and reservoir modeling workflows.

Future research in this line of work could explore modifications to the current setup to allow for the simultaneous training of deep models to track multiple seismic horizons instead of only one horizon, which is one of the key limitations of the present work. This also motivates research into loss functions that can account for the relative ordering of the target horizons and penalize the model for estimating non-geological vertical crossings. Additionally, more studies are required to investigate the behavior of regression-trained models at predicting strongly dipping or discontinuous horizons, horizons interrupted by faults, unconformities etc. Adaptively weighting the geological consistency regularization term to account for underlying geological complexity would be a major research direction in this regard.    

\section{Conclusion}

In this paper, we presented a generic, model-agnostic coordinate regression module for 3D seismic horizon tracking that overcomes the key limitations of standard dense segmentation frameworks. By directly predicting continuous horizon depth coordinates along vertical seismic traces, our approach eliminates the need for artificial label-thickening, spatial upsampling, and heuristic post-processing extraction. Comprehensive evaluations across four distinct backbone architectures ranging from classical convolutional networks (ResNet-50, ResNet-101) to high-capacity modern vision models (ViT-Large, ConvNeXt-Large) demonstrate that replacing segmentation heads with our proposed regression module consistently yields superior quantitative performance and qualitatively smoother horizon picks.

Our findings further highlight three key operational advantages of the proposed formulation: (1) Incorporating an LSTM module across adjacent seismic slices effectively models spatial correlations, enforcing structural coherence along complex geological interfaces. (2) By training exclusively on valid picks and leveraging a geology-based consistency prior, the regression framework avoids the performance degradation caused by false-negative background assumptions inherent to segmentation masks in sparse-pick settings. (3) The model's predictive uncertainty directly reflects underlying geological complexity, providing a reliable confidence metric to automatically highlight ambiguous structural zones for expert geoscientific review.

Ultimately, this work provides a flexible, plug-and-play alternative to dense segmentation for data-driven horizon tracking applications for automated seismic interpretation workflows.

\section{Acknowledgments}

We thank New Zealand Petroleum and Minerals (NZPM) for providing the seismic datasets used in this study. We acknowledge SLB for providing Petrel software to the University of Oklahoma, which supported the interpretation and labeling work conducted in this research. We acknowledge the AASPI consortium for its support and research resources. We also acknowledge the CIVIL lab for providing computational resources through a research credits award from Google Cloud. We thank Dr. Kurt Marfurt and Dr. David Lubo Robles for their insightful discussions and constructive suggestions that contributed to this work.

\appendix
\section{Implementation Details and Hyperparameters}
\label{sec:appendix_implementation}

\subsection{Data Storage, Preparation, and Splits}
\begin{itemize}
    \item \textbf{Data Format:} The 3D seismic volume and ground-truth horizon masks are stored and accessed in Zarr chunked array format.
    \item \textbf{Data Preprocessing:} Seismic amplitudes are standardized globally by subtracting the volume mean $\mu$ and dividing by the standard deviation $\sigma$. Amplitudes are clipped to $[-3\sigma, 3\sigma]$ and scaled to $[-1, 1]$. Single-channel grayscale slices are duplicated across 3 channels ($C=3$) to match pretrained RGB input requirements.
    \item \textbf{Dimensionality:} Spatial dimensions are resized via bilinear interpolation to $H = 224$ and $W = 224$. Target 2D binary horizon masks $M(r, j)$ are mapped to $1\mathrm{D}$ coordinates via $y_j = \min\{r \mid M(r,j)=1\}$ and normalized as $y_j^{\mathrm{norm}} = y_j / H$.
    \item \textbf{Dataset Splitting:} Spatial overlap is minimized by partitioning the volume along the inline axis. Training samples are extracted from inlines 0 through 571 at a stride of 25. Evaluation/testing samples are sampled from inlines 181 through 565 at a stride of 12. Boundary target inlines replicate edge slices to maintain sequence length $T=3$ ($k=1$).
\end{itemize}

\subsection{Architecture and Training Configurations}
\begin{itemize}
    \item \textbf{Backbone Backbone Networks:} Experiments utilize ImageNet-pretrained feature extractors including ResNet-50, ResNet-101, ConvNeXt-Large, and Vision Transformer-Large (ViT-L). The primary benchmark results reported use ConvNeXt-Large with output feature dimension $D = 1536$.
    \item \textbf{Sequence Parameters:} The input context window size is set to $T = 3$ ($k = 1$). The unidirectional LSTM utilizes a hidden dimension $d_h = 256$ with a dropout probability of $0.2$.
    \item \textbf{Optimization Settings:} Models are optimized using Adam with an initial learning rate of $10^{-4}$ and a batch size of 4 for 100 epochs. The regularizing scalar coefficient for smoothness loss is set to $\lambda = 10^{-4}$.
    \item \textbf{Reproducibility:} All experiments are conducted using a fixed global random seed of 42.
\end{itemize}


%








\bibliographystyle{IEEEtran}
\bibliography{bibtex/bib/IEEEexample}



%

\begin{IEEEbiography}[{\includegraphics[width=1in,height=1.25in]{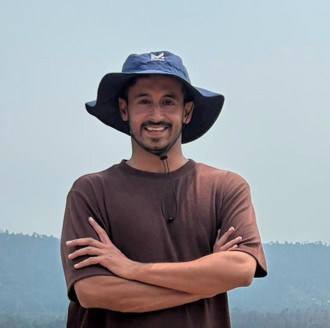}}]{Aniq Ahmad}
is a second-year doctoral student at the University of Oklahoma with a strong interest in the sustainable application of artificial intelligence in the geosciences. He previously worked as a geophysicist at SLB, where he focused on automating interpretation and subsurface workflows. His current research explores how pretrained foundation models and other machine learning approaches can be adapted for seismic interpretation. He serves as secretary of OU's SEG Student Chapter.
\end{IEEEbiography}

\begin{IEEEbiography}[{\includegraphics[width=1in,height=1.25in]{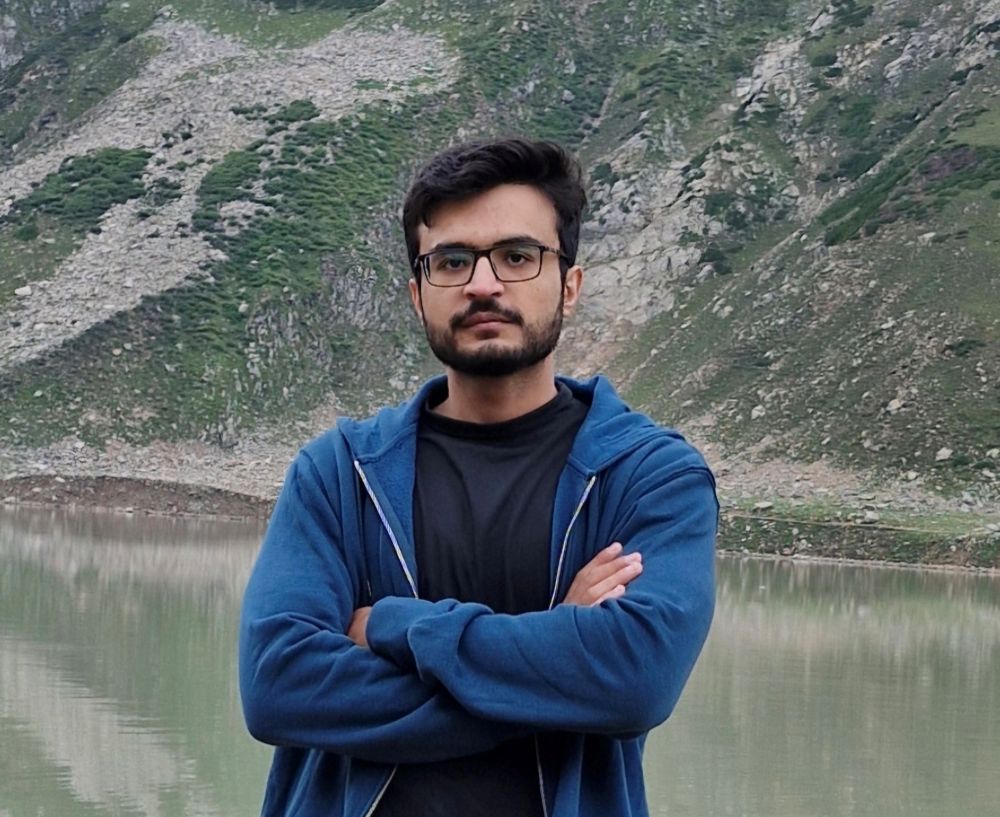}}]{Musham Ahmad Malik}
is currently pursuing the M.S. degree in data science at the Information Technology University (ITU), Lahore, Pakistan, where his research focuses on deep learning for seismic horizon tracking. He is also a Data Scientist with BBI Tech, Lahore. His research interests include computer vision, deep learning for subsurface imaging, and LLM-based agentic systems.
\end{IEEEbiography}

\begin{IEEEbiography}[{\includegraphics[width=1in,height=1.25in]{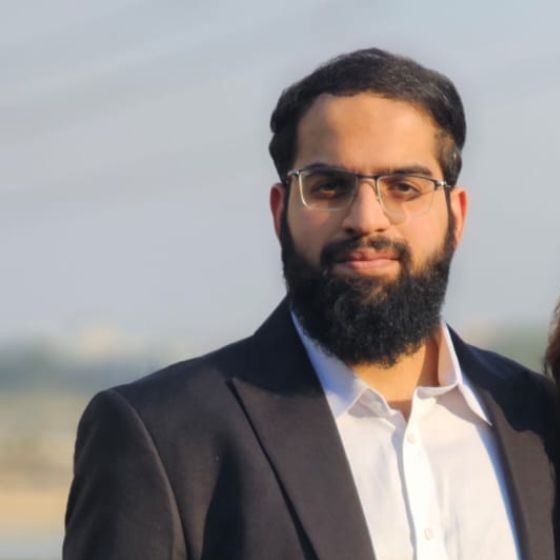}}]{Dr. Ahmad Mustafa (Member, IEEE)}
 is currently an AI Research Fellow with the SDAIA Joint Research Center for AI (JRCAI) at the King Fahd University of Petroleum and Minerals, Saudi Arabia. He was previously an Assistant Professor at the Department of Computer Science based in the Information Technology University (ITU), Lahore, Pakistan. He is the Director of Computational Imaging, Vision, Intelligence and Learning (CIVIL) lab. His research interests span machine learning, weakly supervised learning, and computational image interpretation in scientific computing domains. He obtained his PhD in Electrical and Computer Engineering at the Georgia Institute of Technology, USA. Prior to joining ITU, he worked at Occidental Petroleum, leading the development of large-scale computational methods to streamline subsurface applications.
\end{IEEEbiography}

\begin{IEEEbiography}[{\includegraphics[width=1in,height=1.25in]{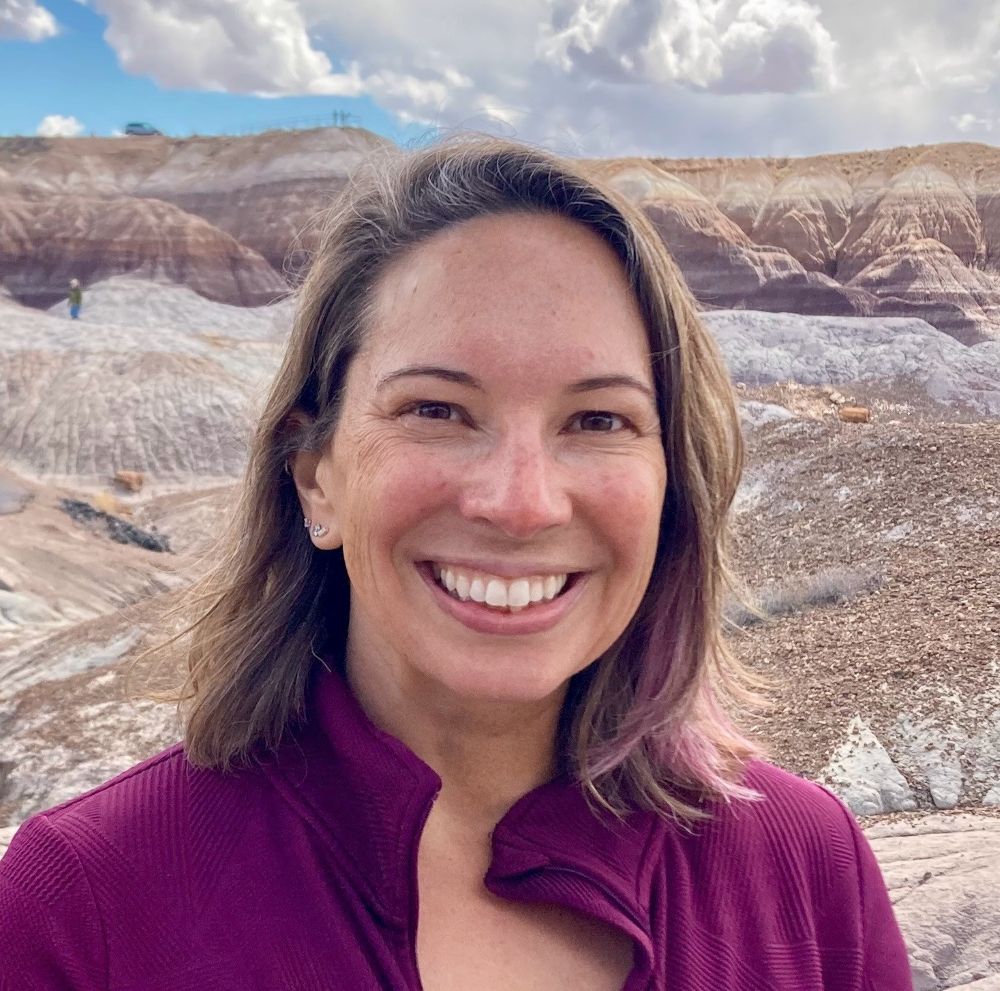}}]{Dr. Heather Bedle}
is the director of the Sustainable Energy Systems Program, and associate professor, Edith Kinney Gaylord Presidential Professor, and Lissa and Cy Wagner Professor of Geosciences at the University of Oklahoma. As PI of the Attribute Assisted Seismic Processing \& Interpretation (AASPI) Consortium, she is passionate about mentoring aspiring scientists and exploring the complex intersections of earth, energy, and environment. With a Ph.D. from Northwestern University, she combines her industry and academic experiences to lead research groups recognized for their interdisciplinary approach, employing methods that revolve around data science and machine learning, and using a variety of data from reflection seismic, earthquake, shallow crustal geophysical methods, and social survey data.
\end{IEEEbiography}








\end{document}